\documentclass{article}

\usepackage{times}
\usepackage{graphicx} % more modern
\usepackage[caption=false]{subfig}

\usepackage{hyperref}
\usepackage{url}
\usepackage{epsfig}
\usepackage{amsmath}
\usepackage{amssymb}

\usepackage{amsthm}
\usepackage{bm}
\usepackage{mathtools}
\usepackage{mathrsfs}
\usepackage{natbib}

\usepackage{algorithm}
\usepackage{algorithmic}
\usepackage{booktabs}
\usepackage{tabu}
\usepackage{multirow}

\usepackage[11pt]{moresize}
\usepackage{array}
\floatstyle{plaintop}
\restylefloat{algorithm}

\newcommand{\thickhline}{%
    \noalign {\ifnum 0=`}\fi \hrule height 1pt
    \futurelet \reserved@a \@xhline
}

\usepackage[accepted]{icml2014}
\pdfoutput=1

\icmltitlerunning{Homophilic Clustering}

\begin{document}

\twocolumn[
\icmltitle{Homophilic Clustering by Locally Asymmetric Geometry}

% unless you've provided the [accepted] option to the icml2014
\icmlauthor{Deli Zhao}{deli\_zhao@htc.com}
\icmladdress{HTC Beijing Advanced Technology and Research Center,
            Beijing, China}
\icmlauthor{Xiaoou Tang}{xtang@ie.cuhk.edu.hk}
\icmladdress{ Department of Information Engineering, The
Chinese University of Hong Kong, Hong Kong}

\icmlkeywords{clustering, graph-based learning, complex networks}

\vskip 0.3in
]

\begin{abstract}
Clustering is indispensable for data analysis in many scientific disciplines. Detecting clusters from heavy noise remains challenging, particularly for high-dimensional sparse data. Based on graph-theoretic framework, the present paper proposes a novel algorithm to address this issue. The locally asymmetric geometries of neighborhoods between data points result in a directed similarity graph to model the structural connectivity of data points. Performing similarity propagation on this directed graph simply by its adjacency matrix powers leads to an interesting discovery, in the sense that if the in-degrees are ordered by the corresponding sorted out-degrees, they will be self-organized to be homophilic layers according to the different distributions of cluster densities,  which is dubbed the Homophilic In-degree figure (the HI figure). With the HI figure, we can easily single out all cores of clusters, identify the boundary between cluster and noise, and visualize the intrinsic structures of clusters. Based on the in-degree homophily, we also develop a simple efficient algorithm of linear space complexity to cluster noisy data. Extensive experiments on toy and real-world scientific data validate the effectiveness of our algorithms. %\vspace{-0.2cm}
\end{abstract}

\section{Introduction}%\vspace{-0.2cm}

Clustering is a fundamental task in machine learning, computer vision, information retrieval, and data mining. Data generated in practice generally possess the property of local or global aggregation in sample spaces due to pattern correlations; this lays the foundation of detecting clusters in data points.

The classic algorithms of clustering are k-means and the hierarchical agglomerative algorithms based on linkages, such as the single, average, and complete linkages. The k-means algorithm iteratively optimizes clusters by minimizing distance squares between the center of each cluster and associated cluster members, which is simple and easily usable. The distance-based partitional method is put forward by the algorithm of Affinity Propagation (AP) \cite{Frey07} that is proven to be fast only with simple manipulations of sparse networks. The most popular linkage algorithms is the average linkage, which measures the structural proximity of pairwise clusters by the arithmetic mean of distances between all members in the two clusters. The framework of hierarchical agglomerative clustering is also applied in the advanced graph-theoretic models, such as graph cycles \cite{Zhao08} and directed linkages \cite{Zhang12}.

In addition to the above conventional frameworks, spectral clustering is a different type of approaches that can cluster data of complex structures. For example, the Normalized Cuts (NCuts) method hierarchically splits data by the graph Laplacian in the divisive way \cite{Shi00}. Alternatively, the k-means or other types of clustering algorithms can be performed on spectral coordinates derived by eigenvectors of graph-Laplacian matrix \cite{Ng01,Meila07,Zhou05}. Spectral embeddings of Laplacian can unfold the underlying manifolds in low-dimensional spaces \cite{Belkin03}. Therefore, spectral clustering is free from the limitation of data structures or distributions. A variant of requiring low-dimensional coordinates was presented by \cite{Lin10}, which is based on matrix power iterations. However, these algorithms will encounter difficulty when clustering data contaminated with noise. Another type of clustering algorithms that have been proven to be noise-robust are based on the application of graph kernels or their analogues. The entries of graph kernel matrices can be viewed as the measurement of global similarities between data points. With the kernel-enhanced similarities, the proximal correlations between data points can be more accurately measured. The hierarchical clustering algorithms are usually employed on graph kernels, such as the matrix power kernel \cite{Newman03,MCL00}, the von Neumann kernel \cite{Katz53}, and the diffusion kernel \cite{Kandola03}. The obvious limitation of such algorithms is that the space complexity is square in the number of data points.

We are now in the era of data deluge. The large scales of data incur two difficulties for data clustering. Firstly, the space complexity of algorithms should be sufficiently low that the available RAM is adequate to run the algorithms. Secondly, the large-scale data usually contains noise or outliers, thereby requiring the algorithms to identify outliers and noise. These issues necessitate the development of clustering algorithms that are robust to noise or outliers with low space complexity.

In this paper, we propose new algorithms to address the issue of accurately clustering noisy data with low complexities of space and time. Our algorithmic framework is based on an intriguing property of directed graphs drawn from data. The asymmetries of the local neighborhoods of each data point lead to a directed graph that is embedded in high-dimensional space. We discover that the arrangement of high-order in-degrees ranked by corresponding sorted out-degrees on such directed graphs breeds the homophilic distribution of data points according to different densities. This density homophily classifies data points into transparent layers according to the values of in-degrees. Noisy data or outliers have low densities such that they are aggregated to form the weakest layer, making it easy to find the boundary between clusters and noise. In addition, the cores of all clusters can be singled out simultaneously by the ratios of in-degrees to out-degrees, thereby greatly facilitating the performance of clustering noisy data. Based on density homophily, we develop a simple algorithm for clustering. Our algorithm passes similarities with local connectivity of directed graphs according to homophilic priority. It attains the better accuracy of clustering with the linear space complexity while maintaining the low time complexity.%\vspace{-0.2cm}

\section{Density Homophily with Digraph} \label{local-cluster}%\vspace{-0.2cm}

Homophily is a concept that describes the behavioral preference of individuals with others who have similar attributes to themselves in social sciences. These attributes include age, gender, race, belief, interests, etc. The conventional work in the seminal paper \cite{McPherson01} and a recent one \cite{Kossinets09} comprehensively studied the homophilic organization in social networks. It has been recently reported that besides popularity modeled by power laws of degree distributions \cite{Barabasi99}, homophily is another dimension of characterizing the preferential attachment of new links in real evolving networks \cite{Papadopoulos12}. We find that high-order in-degrees in geometric digraph show the transparent homophilic distributions with similarity propagation. These homophilic distributions are associated with different densities of clusters in data points.%\vspace{-0.2cm}

\subsection{Neighborhood Asymmetry and Digraph Model}%\vspace{-0.2cm}

 Suppose that a data set $\mathcal{X}=\{\boldsymbol{x}_i|\boldsymbol{x}_i\in \mathcal{M}^d,i=1,\dots,n.\}$ is provided, where $\mathcal{M}^d$ is the $d$-dimensional sample space and $n$ is the number of data points. $\mathcal{M}^d$ may be the mixture of manifolds or multivariate Gaussians. For an arbitrary data point $\boldsymbol{x}_i$, we may search its $k$ nearest neighbors (NNs) $\mathcal{N}_i=\{\boldsymbol{x}_{i_p}|\boldsymbol{x}_{i_p}\in \mathcal{M}^d,p=1,\dots,k.\}$ with respect to a pre-defined distance metric. Assume that another data point $\boldsymbol{x}_j$ is one of NNs of $\boldsymbol{x}_i$, say, $\boldsymbol{x}_j\in\mathcal{N}_i$. For the set $\mathcal{N}_j$ of NNs of $\boldsymbol{x}_j$, there are two possible cases for the structural relationship of $\boldsymbol{x}_i$ to $\boldsymbol{x}_j$: $\boldsymbol{x}_i\in\mathcal{N}_j$ or $\boldsymbol{x}_i\not\in\mathcal{N}_j$. In other words, $\boldsymbol{x}_i$ may not necessarily be the NN of $\boldsymbol{x}_j$ if $\boldsymbol{x}_j$ is the NN of $\boldsymbol{x}_i$. This neighborhood asymmetry is the most fundamental fact of spatial adjacency on local neighborhoods of data cloud. If we locally connect data points by a graph $\mathcal{G}$ of NNs, a weighted adjacency matrix can be formed by%\vspace{-0.1cm}
 \begin{equation}
W_{i\rightarrow j}=
\begin{cases}
%\frac{\mbox{exp}\left(-d^2_{ij}/{\sigma}^2 \right)}{\sum^n_{i=1}\mbox{exp}\left(-d^2_{ij}/{\sigma}^2 \right)},  & \mbox{if } \boldsymbol{x}_i \in  %\mathcal{N}_i \\
\mbox{\textsf{sim}(}\boldsymbol{x}_i,\boldsymbol{x}_j \mbox{)},  & \mbox{if } \boldsymbol{x}_j \in  \mathcal{N}_i \\
1, & \mbox{if } j = i\\
0, & \mbox{ otherwise }
\end{cases},%\vspace{-0.1cm}
\end{equation}
where $i$ successively goes from $1$ to $n$, meaning that the graph $\mathcal{G}$ is constructed row by row. $\mbox{\textsf{sim}(}\boldsymbol{x}_i,\boldsymbol{x}_j \mbox{)}$ presents the value of similarity between $\boldsymbol{x}_i$ and $\boldsymbol{x}_j$, and $\mbox{\textsf{sim}(}\boldsymbol{x}_i,\boldsymbol{x}_j \mbox{)}\in [0,1]$. The similarity measurements may be the cosine value, the distance exponential such as $e^{(-d^2_{ij}/{\sigma}^2)}$, where $\sigma$ is a free parameter, or other variants of similarity measures. $W_{i\rightarrow j}$ is the $(i,j)$-th entry of the weighted adjacent matrix $\boldsymbol{W}$ of graph $\mathcal{G}$. Here we use $i\rightarrow j$ to emphasize that the link from $\boldsymbol{x}_i$ to $\boldsymbol{x}_j$ is directed, thereby forming a digraph $\mathcal{G}$.

Thus for an arbitrary node $i$, there are two types of structural measures: out-degree and in-degree. The out-degree is the sum of weights of out-going links from node $i$ to its neighbors and the in-degree is the sum of weights of in-coming links from neighbors pointing to node $i$. In matrix form, the out-degree vector $\boldsymbol{d}^{out}$ of all nodes can be written as $\boldsymbol{d}^{out}=\boldsymbol{W}\boldsymbol{1}$, where $\boldsymbol{1}$ is the all-one vector of length $n$, and $\boldsymbol{d}^{in}=\boldsymbol{W}^{\top}\boldsymbol{1}$, where $\top$ denotes the matrix transpose. It suffices to note that each node in $\mathcal{G}$ is imposed with a loop of weight $1$, thereby excluding the case of the vanishing in-degrees and out-degrees. The out-degrees and in-degrees are the most elementary ingredients in the characteristics of complex networks.%\vspace{-0.2cm}

\subsection{Similarity Propagation}%\vspace{-0.2cm}

From the viewpoint of paths in $\mathcal{G}$, the structural connectivity modeled by $\boldsymbol{W}$ can be regarded as the linkage of paths of length $1$. Many studies have verified that long paths are favorable for modeling complex structures. For instance, the shortest paths are applied to characterize manifold and network structures \cite{Tenenbaum00}. Long cycles can convey the high-level information of balance in signed networks \cite{Zhao08,Chiang11}. With long paths, the membership affinities within an arbitrary network community can be more accurately enhanced \cite{Katz53,Newman03}. Similarity propagation by walks is the simplest and most intuitive of the various applications of paths. It can be written simply by matrix power $\boldsymbol{W}^t$, where $t$ presents the length of paths to be investigated. The $(i,j)$-th entry $W^t_{i\rightarrow j}$ of $\boldsymbol{W}^t$ can be interpreted as a kind of accumulative similarity between $\boldsymbol{x}_i$ and $\boldsymbol{x}_j$ by passing similarities in digraph $\mathcal{G}$ by $t$ steps. To make it clear, we expand $W^t_{i\rightarrow j}$ by graph representation to give %\vspace{-0.1cm}
 \begin{equation}\label{eq-similarity}
W^t_{i\rightarrow j}=\sum_{\{\mbox{{\ssmall all possible paths of length }} t\}}\prod_{\{\mbox{{\ssmall one of paths of length }} t\}}W_{i\rightarrow j}.%\vspace{-0.1cm}
\end{equation}
From (\ref{eq-similarity}), it is easy to see that $W^t_{i\rightarrow j}$ is essentially a global sum-product similarity generated by all possible paths of length $t$ that connect node $i$ and node $j$. The path-based similarity can capture the structural correlation of deep connections between data points. If we regard each data point as a human individual and the whole data set as the society, the growth of $\boldsymbol{W}^t$ can be viewed as the dynamic process of individual social interactions. Therefore, we can apply the social concepts and principles for data analysis. The homophily pertaining to data we underscore is such a social property of data points.

With $\boldsymbol{W}^t$, we can define the $t$-order degrees as $\boldsymbol{d}^{in}=\boldsymbol{W}^t\boldsymbol{1}$ and $\boldsymbol{d}^{out}=(\boldsymbol{W}^t)^{\top}\boldsymbol{1}$. For convenience of representation, we have omitted the scripts of `$t$' in $\boldsymbol{d}^{in}$ and $\boldsymbol{d}^{out}$. It may be inferred from context. It is computationally prohibitive to directly compute $\boldsymbol{W}^t$ for a large $n$, because $\boldsymbol{W}^t$ turns to be a fully dense matrix for a moderate $t$. Actually, $\boldsymbol{d}^{in}$ and $\boldsymbol{d}^{out}$ can be derived by sparse-matrix-vector products iteratively. To maintain the values of $\boldsymbol{d}^{in}$ and $\boldsymbol{d}^{out}$, we perform the sum-to-one normalization during iteration. The procedures are provided in Algorithm 1.
%
%\begin{algorithm}[htb]
%\caption{\textbf{Compute $t$-order out-degrees and in-degrees}}\vspace{-0.3cm}
%\label{alg:degree}
%\begin{algorithmic}[1]
%\REQUIRE
%The weighted adjacency matrix $\boldsymbol{W}$ and the free parameter $t$.
%\STATE Initialization. $\boldsymbol{d}^{in}\leftarrow \boldsymbol{1}$ and $\boldsymbol{d}^{out}\leftarrow \boldsymbol{1}$.  %\label{code:fram:extract}
%%
%\FOR{$tt=1$ to $t$}
%\STATE $d^{in}_i\leftarrow \sum^n_{j=1}W_{ij}d^{in}_j$ and $d^{out}_j\leftarrow \sum^n_{i=1}W_{ij}d^{out}_i$.
%\STATE $a = \frac{1}{2}\sum^n_{i=1}(d^{in}_i+d^{out}_i)$. $d^{in}_i\leftarrow d^{in}_i/a$ and $d^{out}_i\leftarrow d^{out}_i/a$.
%\ENDFOR
%\label{code:fram:add}
%\ENSURE $\boldsymbol{d}^{in}$ and $\boldsymbol{d}^{out}$.
%\end{algorithmic}\vspace{-0.2cm}
%\end{algorithm}
%%
Notice that we use the same scale constant to normalize $\boldsymbol{d}^{in}$ and $\boldsymbol{d}^{out}$ in each iteration in Algorithm 1. Such manipulation is crucial for the usage of in-degrees and out-degrees, which will be presented in the following section.%\vspace{-0.2cm}
\begin{table*}[tb]
%\caption{ Algorithms of computing $t$-order dual degrees and plotting the homophilic in-degree figure (HI figure). }\vspace{-0.3cm}
\label{table:algorithms}
%\vskip 0.15in
\begin{center}
%\begin{tabular}{|l|l|}
\begin{tabular}{l|l}
\hline
\textbf{Algorithm 1} $t$-Order Dual Degrees         & \textbf{Algorithm 2} HI Figure\\
\hline
\textbf{Input}: The graph matrix $\boldsymbol{W}$ and the integer $t$.    & \textbf{Input}: The $t$-order $\boldsymbol{d}^{in}$ and $\boldsymbol{d}^{out}$.\\
1: Initialization. $\boldsymbol{d}^{in}\leftarrow \boldsymbol{1}$ and $\boldsymbol{d}^{out}\leftarrow \boldsymbol{1}$.                                     & ~~~~~~~~~~~~The index vector $\boldsymbol{s}=[1,\dots,n]$.\\
2: \textbf{for} {$tt=1$ \textbf{to} $t$}      & 1: $\vec{\boldsymbol{d}}^{out}\leftarrow \mbox{\textsf{sort}}(\boldsymbol{d}^{out})$ in descending order: \\
3:~~~~~~$\boldsymbol{d}^{out}\leftarrow \boldsymbol{W}\boldsymbol{d}^{out}$  and  $\boldsymbol{d}^{in}\leftarrow \boldsymbol{W}^{\top}\boldsymbol{d}^{in}$.                         &~~~~Record the associated index order $\vec{\boldsymbol{s}}$. \\
4:~~~~~~ $a = \frac{1}{2}(\boldsymbol{d}^{in}+\boldsymbol{d}^{out})^{\top}\boldsymbol{1}$.       &  2: Order $\boldsymbol{d}^{in}$ by $\vec{\boldsymbol{s}}$, $\vec{\boldsymbol{d}}^{in}\leftarrow \boldsymbol{d}^{in}(\vec{\boldsymbol{s}})$. \\
5:~~~~~~$\boldsymbol{d}^{in}\leftarrow \boldsymbol{d}^{in}/a$  and  $\boldsymbol{d}^{out}\leftarrow \boldsymbol{d}^{out}/a$.        &\textbf{Output}: Figure: \textsf{plot}($\boldsymbol{s}$, $\vec{\boldsymbol{d}}^{out}$) ; \textsf{plot}($\boldsymbol{s}$, $\vec{\boldsymbol{d}}^{in}$).\\
6: \textbf{end}  & \\
\textbf{Output}: $\boldsymbol{d}^{in}$ and $\boldsymbol{d}^{out}$.                                                     &  \\
\hline
\end{tabular}
\end{center}
%\vskip -0.1in
\end{table*}

\subsection{In-degree Homophily}%\vspace{-0.2cm}

An interesting property of the $t$-order $\boldsymbol{d}^{in}$ and $\boldsymbol{d}^{out}$ is that the in-degrees are self-organized to be homophilic layers if ordered by the associated out-degrees. The in-degrees reflect the popularity of nodes in digraph $\mathcal{G}$ \cite{Barabasi99,Papadopoulos12}, thereby differentiating the cluster densities of data points. In this way, the density distribution of clusters may be accurately visualized, providing a powerful avenue for intuitively analyzing clusters. We present the specific steps of illustrating the homophilic in-degrees (HI) in Algorithm 2. For simplicity, we call the visualization the HI figure. %\vspace{-0.2cm}
%%
%\begin{algorithm}[htb]
%\caption{\textbf{Homophilic In-degree figure} (HI figure)}\vspace{-0.3cm}
%\label{alg:homophily}
%\begin{algorithmic}[1]
%\REQUIRE The $t$-order $\boldsymbol{d}^{in}$ and $\boldsymbol{d}^{out}$. The index vector $\boldsymbol{s}=[1,\dots,n]$.
%\STATE Sort $\boldsymbol{d}^{out}$ in descending order: $\vec{\boldsymbol{d}}^{out}\leftarrow \mbox{\textsf{sort}}(\boldsymbol{d}^{out})$ and record %the associated index order $\vec{\boldsymbol{s}}$.
%\STATE Order $\boldsymbol{d}^{in}$ according to $\vec{\boldsymbol{s}}$: $\vec{\boldsymbol{d}}^{in}\leftarrow %\boldsymbol{d}^{in}(\vec{\boldsymbol{s}})$.
%\ENSURE Figure: \textsf{plot}($\boldsymbol{s}$, $\vec{\boldsymbol{d}}^{in}$); \textsf{plot}($\boldsymbol{s}$, $\vec{\boldsymbol{d}}^{out}$).
%\end{algorithmic} \vspace{-0.2cm}
%\end{algorithm}
%%

\begin{figure*}[t]
%\vskip 0.2in
\begin{center}
\begin{tabular}{cccc}
   \includegraphics[scale=0.48]{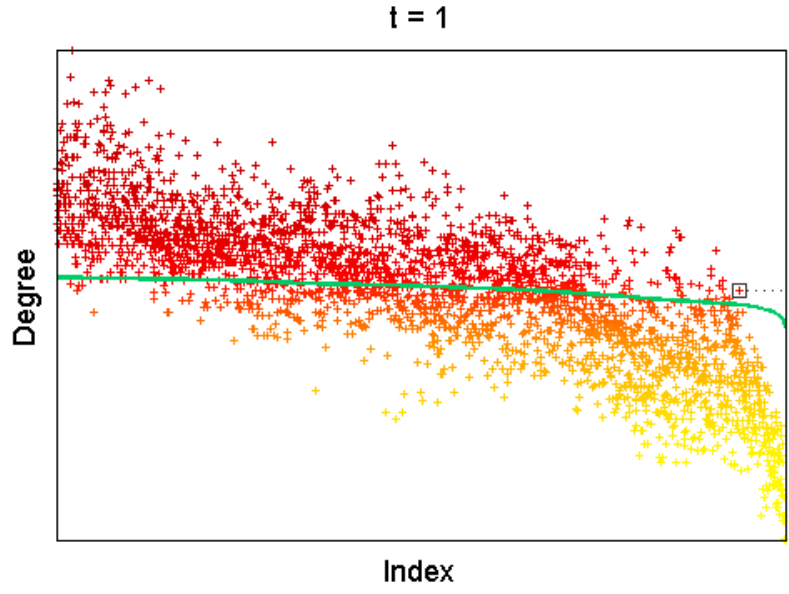}&
   \includegraphics[scale=0.48]{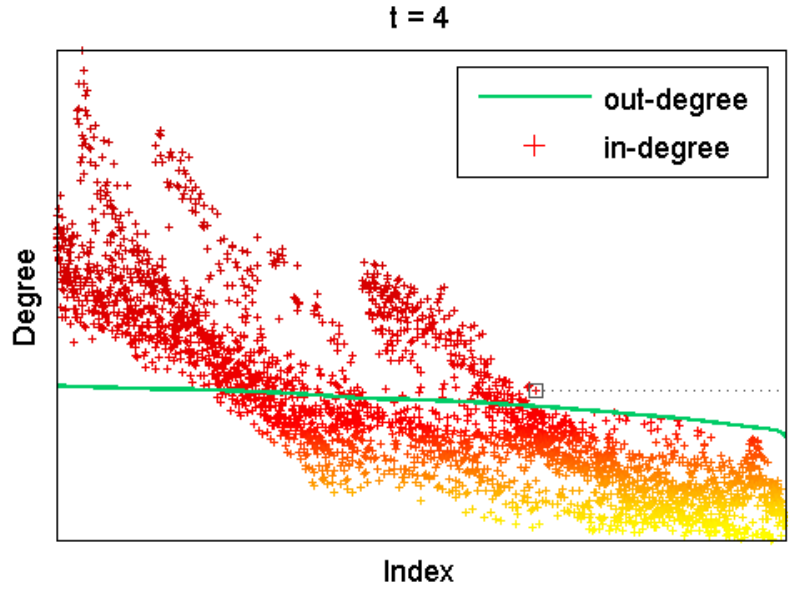}&
   \includegraphics[scale=0.48]{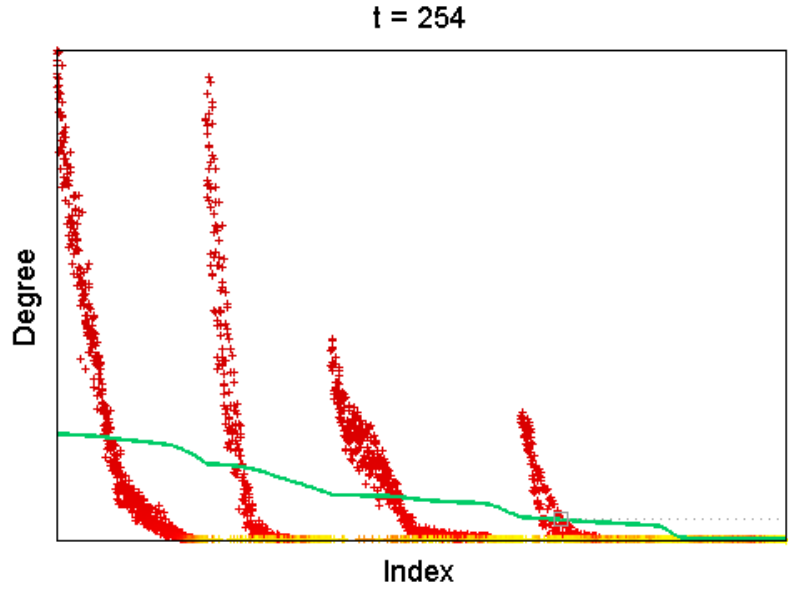}&
   \includegraphics[scale=0.48]{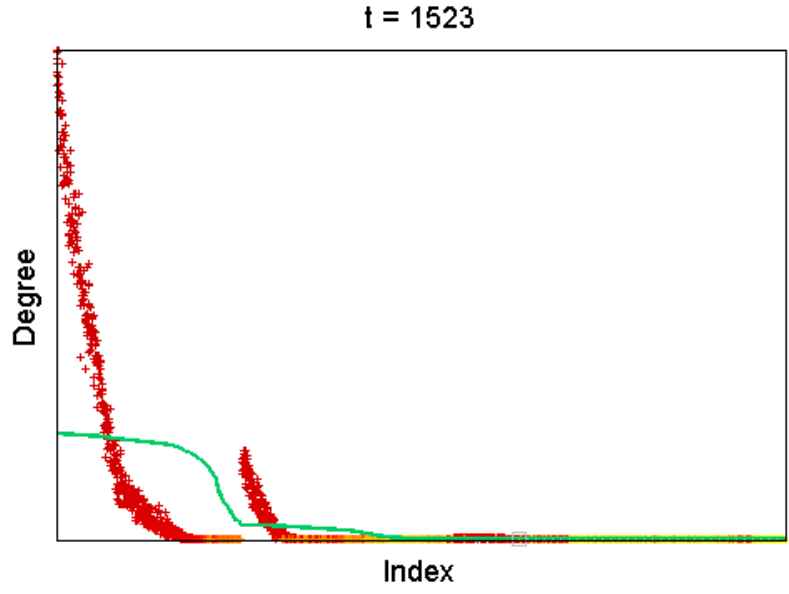}\\
   \includegraphics[scale=0.46]{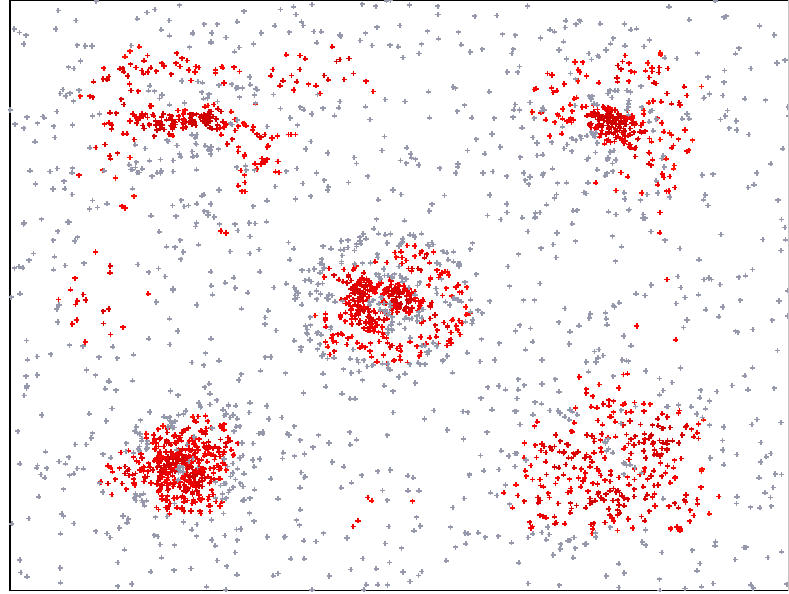}&
   \includegraphics[scale=0.46]{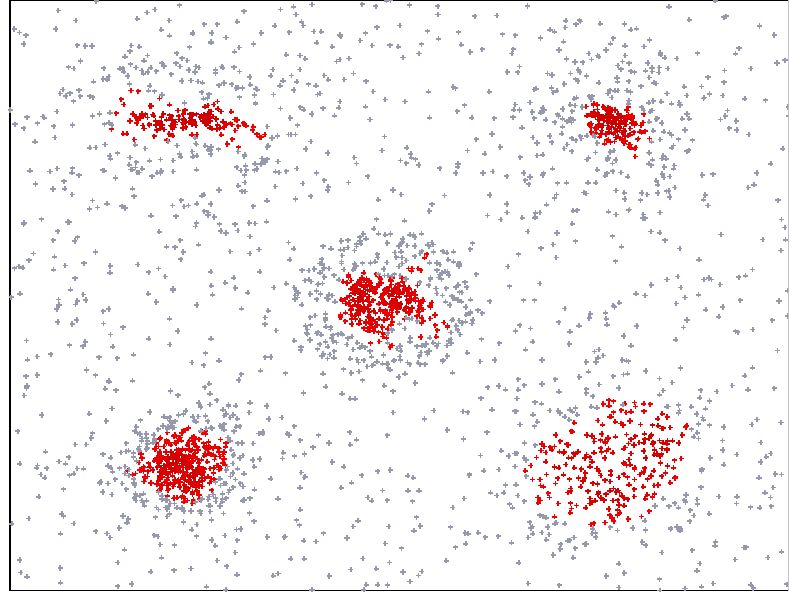}&
   \includegraphics[scale=0.46]{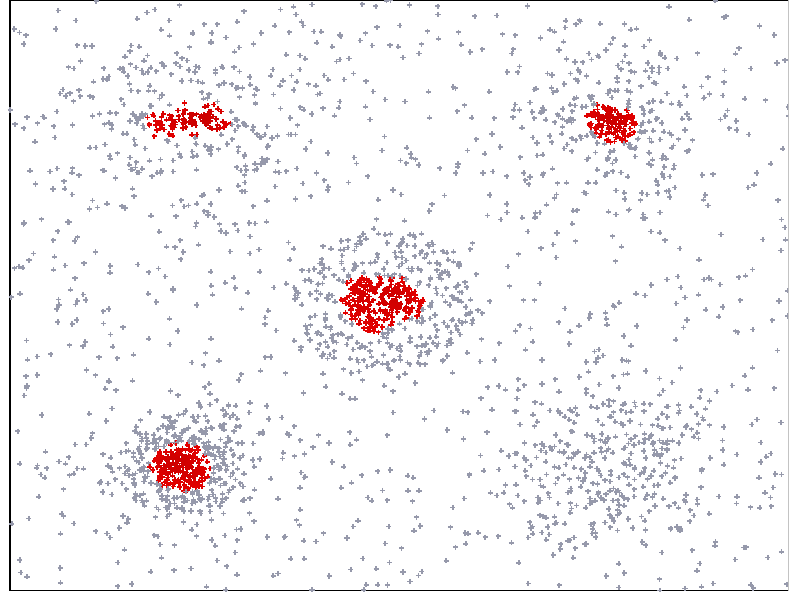}&
   \includegraphics[scale=0.46]{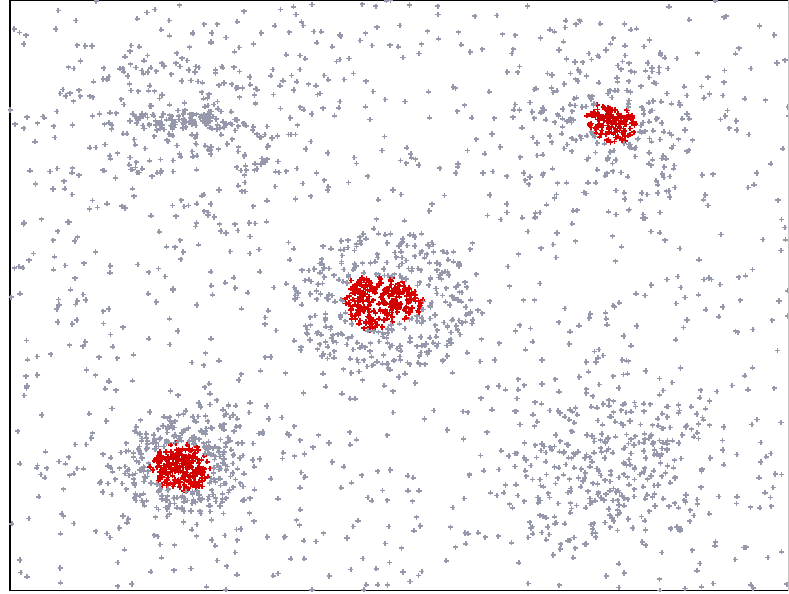}\\
   (a) $t=1$. & (b) $t=4$. & (c) $t=254$. & (d) $t=2909$.\\
   \end{tabular}
\end{center}%\vspace{-0.1in}
   \caption{Toy example of the homophilic in-degree figure (HI figure). The 2D data contain five clusters of different densities in heavy noise, and the five clusters are nested with small clusters. Figures in the first row are the HI figures, while those in the second row are the corresponding cluster cores. The values of $t$ are selected from the jump transitions in Figure~\ref{fig:toyDistanceMean} (a).}%\vspace{-0.5cm}
\label{fig:layerCore}
%\vskip -0.05in
\end{figure*}

Examples of the HI figure on toy data are shown in the first row of Figure~\ref{fig:layerCore}. We can see that there is no clear regular orderliness for $t=1$, which is the case that is most frequently adopted in the analysis of networks. However, the transparent in-degree layers gradually emerge when $t$ goes large, and the difference of altitudes of layers become significant with $t$ approaching $n$. We colorize the HI figure according to clusters and noise, as shown in Figures~\ref{fig:toyClusterLayer} (d) and (g). It is clear that the higher the density of cluster, the nearer the associated in-degree layer approaches the $y$ axis. Moreover, these in-degree layers differentiate according to the distribution of cluster densities. We call the phenomenon of aggregation of in-degrees as the in-degree homophily. For much clearer illustration, we present the complete process of deformation of in-degree homophily in Video 1 of Supplementary Material\footnote{All the supplementary materials of this paper are available at http://sites.google.com/site/zhaodeli/}.

Interestingly, the homophily in social science was vividly described as ``birds of a feather flock together" \cite{McPherson01}. For the geometric network here, we can clearly observe that the overall shape of the in-degree homophily in the HI figure is really like the wing of a bird. Refer also to Figure~\ref{fig:toyClusterLayer} (e) for a better example.%\vspace{-0.2cm}
%-----------------------------------
%-----------------------------------

\section{Homophilic Clustering}%\vspace{-0.2cm}
\label{ALE}
We can develop useful methods for clustering with the interesting characteristic of the in-degree homophily, including extraction of cluster cores, detection of cluster-to-noise boundary, and algorithms of clustering with low price.%\vspace{-0.2cm}

\subsection{Cluster-Core Extraction}%\vspace{-0.2cm}
Clustering will be easy if we can accurately locate the core of each cluster. With the HI figure, this problem becomes simple to handle. We single out the data points whose $t$-order in-degrees are larger than the corresponding out-degrees. If seen from the HI figure, these are data points whose in-degrees lie above the out-degree curve. As Figures~\ref{fig:layerCore} (a)-(d) show, the separated points exactly consist of the cores of clusters for a moderate $t$. It attains messy results in the case of $t=1$. The extracted cores become well-shaped with the growth of in-degree homophily over $t$. This critical clue leads us to define the homophilic coefficient for each node by%\vspace{-0.1cm}
%
%$\hbar=\frac{d^{in}_t}{d^{out}_t}$
\begin{equation}
{\hbar}_i=\frac{d^{in}_i}{d^{out}_i}.%\vspace{-0.1cm}
\end{equation}
The homophilic coefficient ${\hbar}_i$ of node $i$ measures the degree that this node aggregates to be a member of a cluster. The larger the homophilic coefficient, the more important the node is from the clustering perspective. Therefore, we take the cluster cores out from noise by using ${\hbar}_i\geq 1$ for a proper $t$.

The homophilic coefficient of order $1$ was previously proposed and applied for the detection of communities in the complex networks of the internet, genes, etc. \cite{Maslov02,Radicchi04}. For our geometric network, however, the $1$-order homophilic coefficient fails to measure the popularity of clusters. However, we can also observe that the weak layers successively decay below the out-degree curve with the increase of $t$, meaning that the cores of density-weak clusters are percolated out. Therefore, we need to formulate rules to attain an applicable $t$.%\vspace{-0.2cm}
%
%-----------------------------------
\begin{figure}[t]
%\vskip 0.2in
\begin{center}
\begin{tabular}{c}
   \includegraphics[scale=0.6]{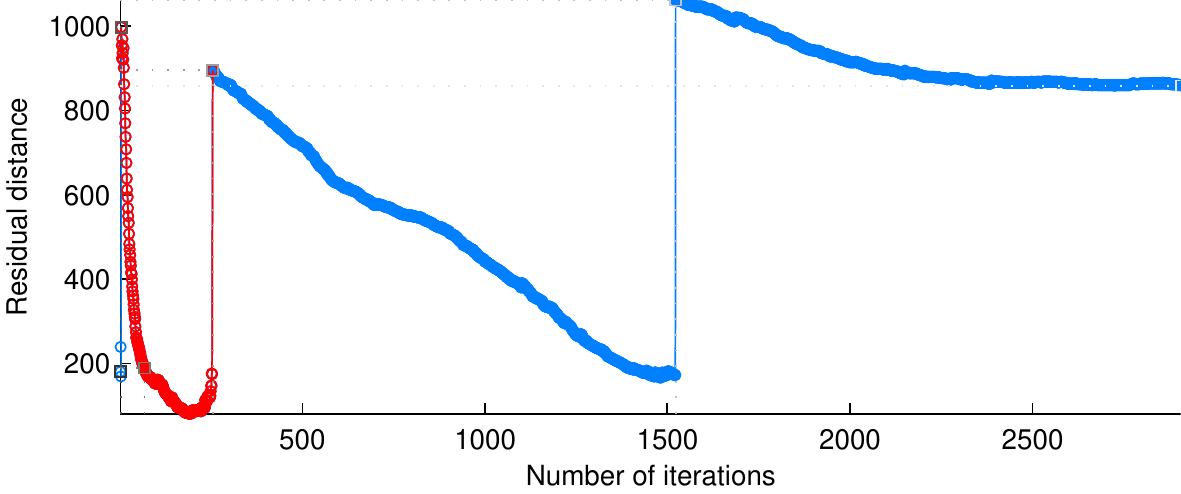}\\
   (a) Residual distance over $t$.\\
   \includegraphics[scale=0.6]{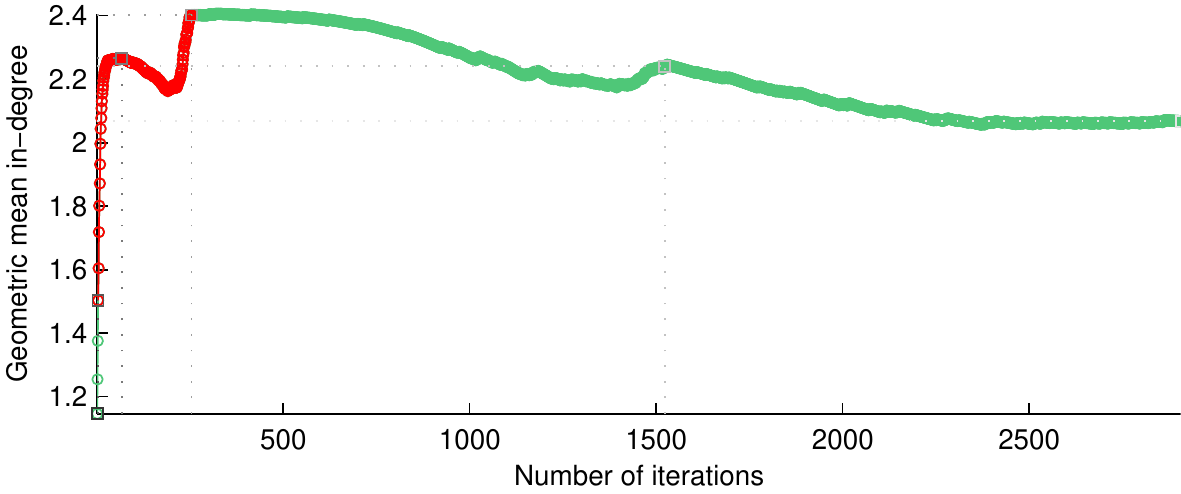}\\
     (b) Geometric mean of $\boldsymbol{d}^{in}$ in cluster cores over $t$.
   \end{tabular}
\end{center}%\vspace{-0.1in}
   \caption{Residual distance and geometric mean of truncated in-degrees. The figures are best viewed in color and large size.}%\vspace{-0.5cm}
\label{fig:toyDistanceMean}
%\vskip -0.05in
\end{figure}
%-----------------------------------

\subsection{Cluster-to-Noise Boundary}%\vspace{-0.2cm}
\label{section:boundary}
The homophilic in-degree layer containing noise is easily identified due to the fact that the connection of network formed by noise is relatively weak. This means that the noise layer always lies at the tail of the HI figure. Another portion of noise consists of the bottom base of the HI figure, which are yielded by connections between clusters and noise. Therefore, the noise layer will be the first one to decay below the out-degree curve. When the noise layer disappears, the second-weakest layer will slither towards the tail of the HI figure over $t$ until the growth of homophily converges. This dynamics of in-degree layers can be clearly observed from Video 1. In this way, we obtain a cue that quantitatively describes the deformation of in-degree layers. To do this, we define the residual distance of the HI figure: it is the distance between the right $y$ axis to the point above the out-degree curve that is vertically nearest to it. We mark the points with gray squares and draw the distance paths with dotted lines in Figure~\ref{fig:toyDistanceMean}. The trajectory of the residual distance over $t$ is depicted in Figure~\ref{fig:toyDistanceMean}, where we see that there is a jump transition when a weak layer decays. Therefore, we extract the largest cores of all clusters at the time at which the last point in the noise layer was just percolated by the out-degree curve. This pivotal time for the toy example is $t=4$ at the first peak of jump transitions. The corresponding HI figure is Figure~\ref{fig:layerCore} (b). We denote the set of largest cores by $\mathcal{C}_{\text{max}}$. To discriminate noise from clusters, we need to guarantee that the local density of any member in $\mathcal{C}_{\text{max}}$ is larger than all the members in $\mathcal{C}_{\text{noise}}$, where $\mathcal{C}_{\text{noise}}$ denotes the set of noise. Thus, we conclude the criterion of identifying the boundary between clusters and noise. Formally, we define the local density ${\eta}_i$ of $\boldsymbol{x}_i$ by the average of similarities in $\mathcal{N}_i$, i.e., ${\eta}_i=\frac{1}{k}\sum^k_{p=1}\mbox{\textsf{sim}(}\boldsymbol{x}_i,\boldsymbol{x}_{i_p} \mbox{)}$.
%
%\begin{equation}
%{\eta}_i=\frac{1}{k}\sum^k_{p=1}\mbox{\textsf{sim}(}\boldsymbol{x}_i,\boldsymbol{x}_{i_p} \mbox{)}.\vspace{-0.1cm}
%\end{equation}
%
Investigating the distance or similarity of $\boldsymbol{x}_i$ to its $k$-th NN is a general way of estimating local density of $\boldsymbol{x}_i$ \cite{Byers98}. Here we use the average to enhance the robustness of estimator. Let%\vspace{-0.1cm}
\begin{equation}
{\eta}^{\text{min}}_{\mathcal{C}_{\text{max}}}=\arg\min_{\boldsymbol{x}_i\in\mathcal{C}_{\text{max}}} {\eta}_i.%\vspace{-0.1cm}
\end{equation}
The set of points in clusters can then be detected by%\vspace{-0.1cm}
\begin{equation}\label{eq:cluster}
\mathcal{C}_{\text{cluster}} = \{\boldsymbol{x}_i| {\eta}_i \geq {\eta}^{\text{min}}_{\mathcal{C}_{\text{max}}}, \boldsymbol{x}_i \in \mathcal{X}\}.%\vspace{-0.1cm}
\end{equation}
The separated clusters from noise shown in Figure~\ref{fig:toyClusterLayer} (c) demonstrate that ${\eta}^{\text{min}}_{\mathcal{C}_{\text{max}}}$ is an effective estimator of cluster-to-noise boundary.

\subsection{Determination of Powers}
\label{section:powers}
\begin{figure*}[t]
%\vskip 0.2in
\begin{center}
\begin{tabular}{cccc}
   \includegraphics[scale=0.45]{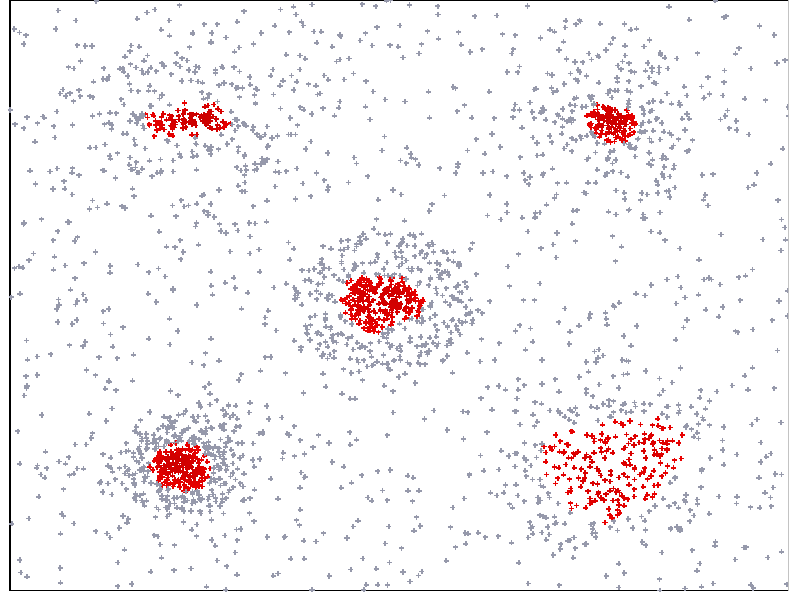}&
   \includegraphics[scale=0.45]{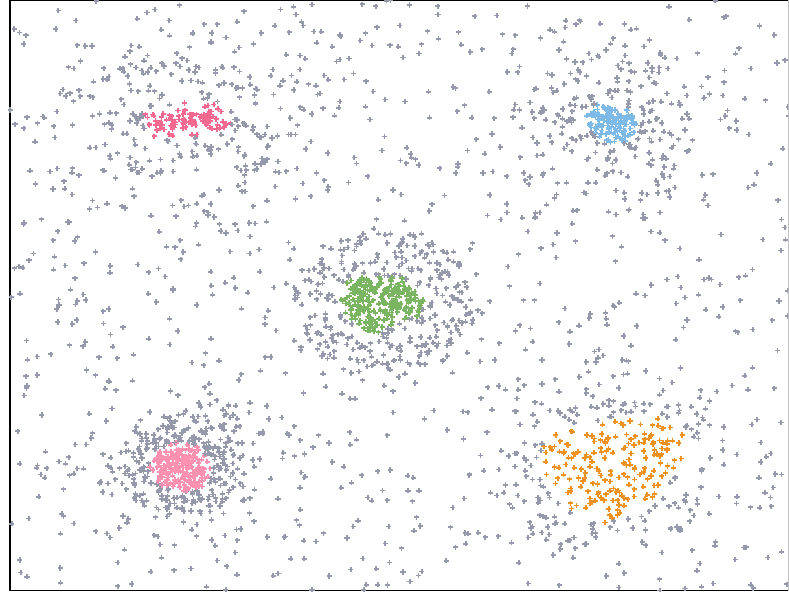}&
   \includegraphics[scale=0.45]{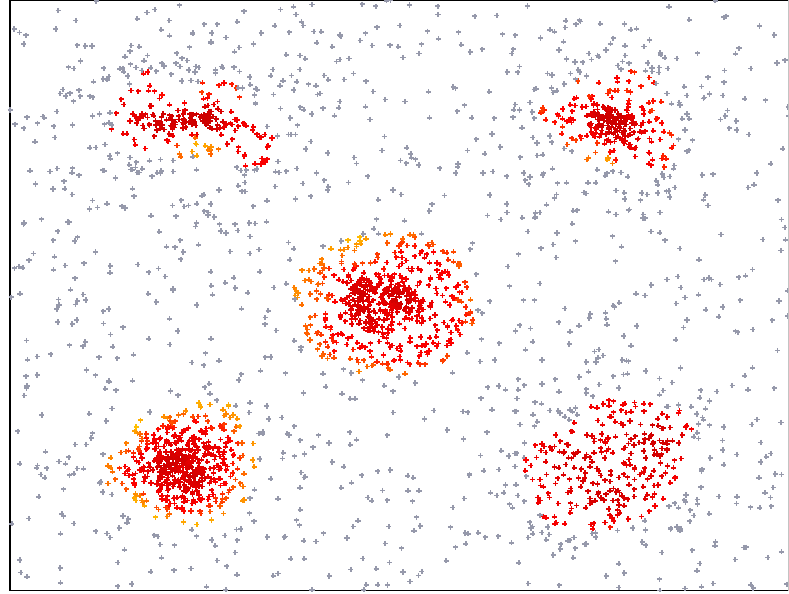}&
   \includegraphics[scale=0.45]{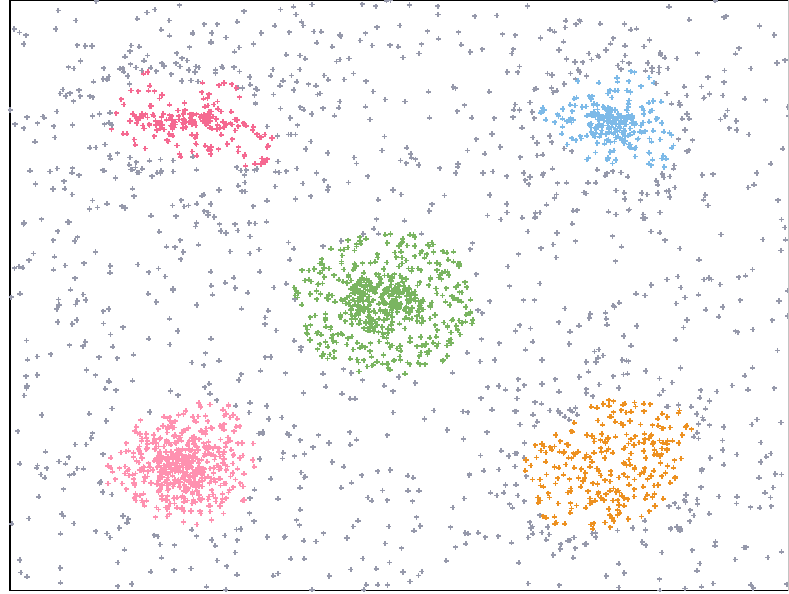}\\
   (a) Optimal cores. & (b) Core clusters. & (c) Cluster set. & (d) Clustering.\\
   \multicolumn{1}{c}{ \includegraphics[width=38mm,height=35mm]{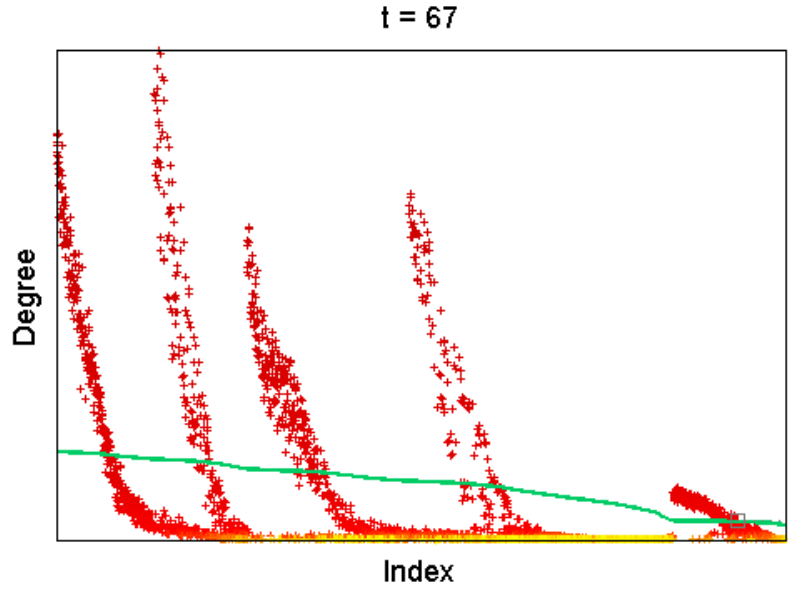} }&
    %\multicolumn{1}{c}{ \includegraphics[scale=0.4]{figure/toy/toy_layerT67.eps} }&
   \includegraphics[width=40mm,height=35mm]{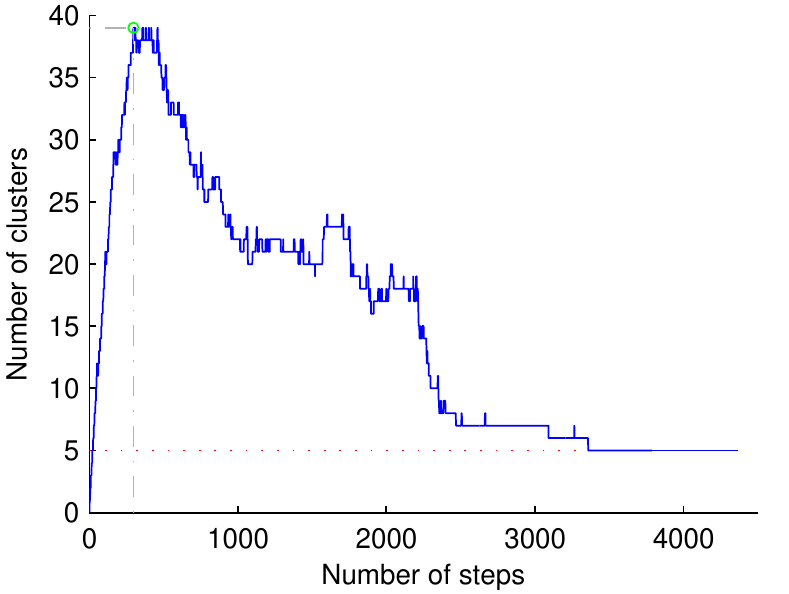}&
   \multicolumn{2}{c}{ \includegraphics[scale=0.42]{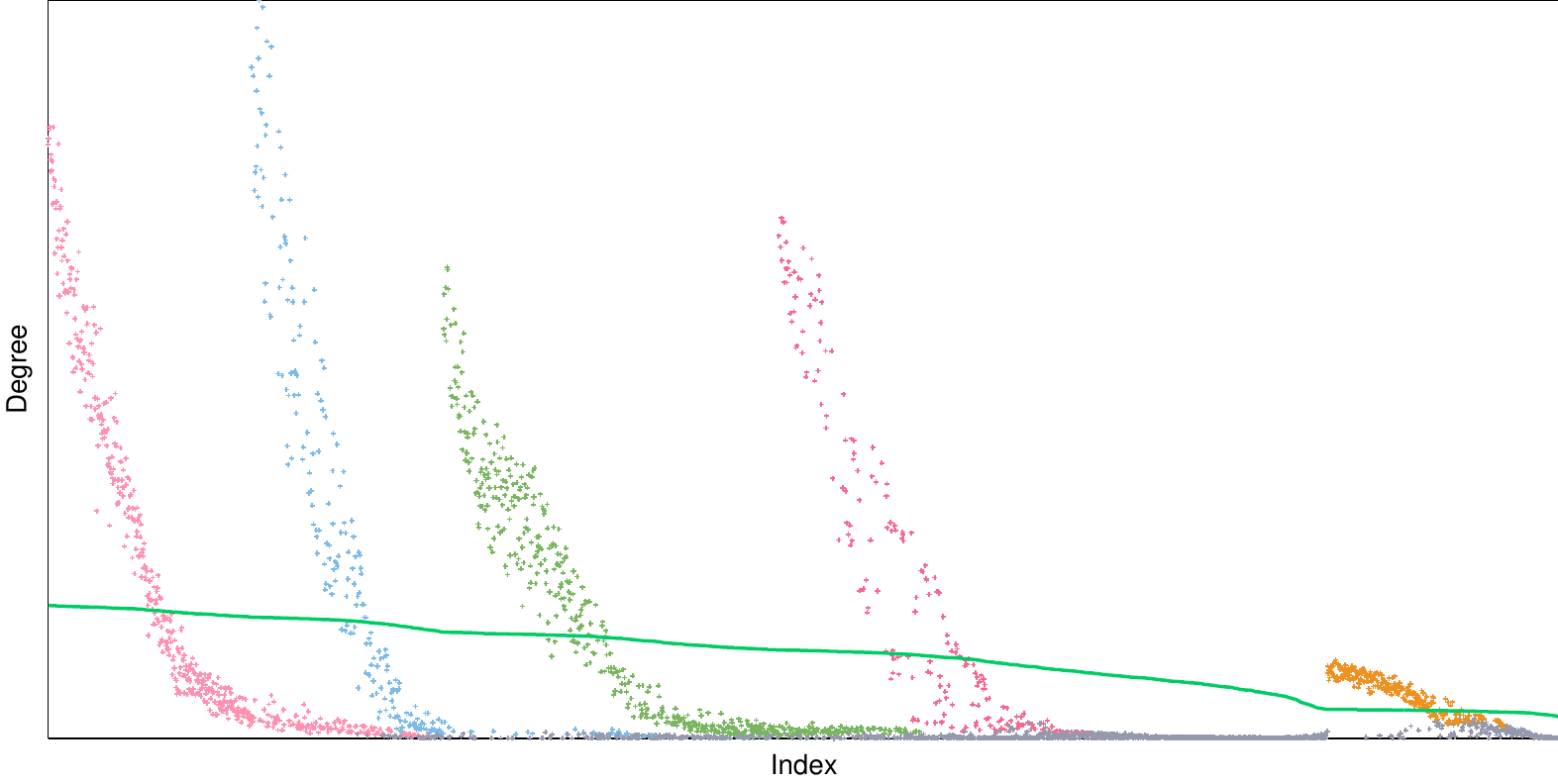} }\\
   (e) Optimal HI figure. & \multicolumn{1}{c}{ (f) Cluster number.}  &  \multicolumn{2}{c}{ (g) Cluster-colorized HI figure. }\\
   %\multicolumn{2}{c}{ (e) }  &  \multicolumn{2}{c}{ (f) }\\
   \end{tabular}
\end{center}%\vspace{-0.1in}
   \caption{Core extraction and clustering. (a) and (b) are the extracted cores and the corresponding HI figure at $t=67$, respectively. This value is determined by the optimal condition in Section~\ref{section:powers}. (b) is the clustering result of cores by homophily-guided mergence. (c) is the complete set of clusters determined by formula~\ref{eq:cluster}. (d) Clustering by attachment in formula~\ref{eq:aggregate}. (f) is the curve of the number of clusters when merging cluster cores. We set $k_c=5$ for toy data. (g) is the optimal HI figure colorized by colors of clusters. These figures are best viewed in color and large size.}%\vspace{-0.5cm}
\label{fig:toyClusterLayer}
%\vskip -0.05in
\end{figure*}

To extract better cluster cores, we must further determine another $t$. The interval between the first and second jump transitions is the feasible set in which the selected $t$ will produce the complete cores, because each cluster has points above the out-degree curve in this interval. We mark the feasible interval of determining $t$ by red circles in Figure~\ref{fig:toyDistanceMean} (a). An optimal $t$ for singling out cores should yield the optimal homophilic layers. Thus, a natural criterion is that the truncated in-degree layers by ${\hbar}_i\geq 1$ is maximally uniform, in the sense that the difference between the truncated in-degree layers of dense clusters and that of sparse clusters is minimized. By this criterion, we can select the balanced cores for all clusters, which is more favorable for clustering. A simple measurement for this optimality is the geometric mean of truncated in-degrees by ${\hbar}_i\geq 1$, showing that %$g_t = \left(\prod_{{\hbar}_i\geq 1}d^{in}_i\right)^{\frac{1}{|\mathcal{C}^t_{\text{core}}|}}$,
\begin{equation}
g_t = \left(\prod_{{\hbar}_i\geq 1}d^{in}_i\right)^{\frac{1}{|\mathcal{C}^t_{\text{core}}|}},%\vspace{-0.1cm}
\end{equation}
where $\mathcal{C}^t_{\text{core}}$ is the set of data points satisfying ${\hbar}_i\geq 1$. Figure~\ref{fig:toyDistanceMean} (b) illustrates the curve of $g_t$. The growth of strong layers and the reduction of weak layers shape the $g_t$ curve with local maxima and minima. An optimal $t$ we expect is at the local maxima in the feasible interval. The selected cores on toy data are shown in Figure~\ref{fig:toyClusterLayer} (a) and the associated HI figure in Figure~\ref{fig:toyClusterLayer} (e).%\vspace{-0.2cm}
%
%-----------------------------------
%-----------------------------------

\subsection{Homophilic Clustering}%\vspace{-0.2cm}
\label{clusteringAlgorithm}
\subsubsection{Pair Merging}
With the extracted clusters and cluster cores, one can develop diverse approaches for clustering. Here we present a simple method based on the homophily-guided mergence of nodal links. Denote the set of extracted cores by $\mathcal{C}^t_{\text{core}}$, where $t$ is optimally determined. For each $\boldsymbol{x}_i\in\mathcal{C}^t_{\text{core}}$, we take its $k_c$ nearest neighbors. Here $k_c$ is a small constant, generally, $k_c\in[1,5]$. For the ideal case, we can directly merge these selected nearest neighbors in $\mathcal{C}^t_{\text{core}}$ to get clusters of cores if they are connected. For complex data, however, there may still be links between cores, which may deteriorate clustering results. To maintain the robustness to noisy links, we define a homophily-weighted similarity $\textsf{hsim}(\boldsymbol{x}_i,\boldsymbol{x}_{j} )$ between selected $k_c$ NNs for cluster merging, giving that%\vspace{-0.2cm}
\begin{equation}
\mbox{\textsf{hsim}(}\boldsymbol{x}_i,\boldsymbol{x}_{i_p} \mbox{)} = \hbar_i\hbar_{i_p}\mbox{\textsf{sim}(}\boldsymbol{x}_i,\boldsymbol{x}_{i_p} \mbox{)},
%\mbox{if } \boldsymbol{x}_i,\boldsymbol{x}_{i_p} \in\mathcal{C}^t_{\text{core}} \mbox{ and } \boldsymbol{x}_{i_p}\in\mathcal{N}^{k_c}_i, p=1,\dots,k_c.\vspace{-0.1cm}
\end{equation}
if $\boldsymbol{x}_i,\boldsymbol{x}_{i_p} \in\mathcal{C}^t_{\text{core}}$ and $\boldsymbol{x}_{i_p}\in\mathcal{N}^{k_c}_i, p=1,\dots,k_c$. The $\textsf{hsim}(\boldsymbol{x}_i,\boldsymbol{x}_{j} )$ is the pairwise similarity weighted by the homophilic coefficients of associated NN pairs. This constraint ensures that the priority of messages passing is already along paths of the high homophily, thereby making the merging procedure robust to noisy links. With homophilic similarity, we can merge data pairs one by one from the largest $\textsf{hsim}(\boldsymbol{x}_i,\boldsymbol{x}_{j} )$ to the smallest one if they share mutual links, until the procedure converges or a given number of clusters is identified.

Figure~\ref{fig:toyClusterLayer} (b) shows the detected clusters of cores and Figure~\ref{fig:toyClusterLayer} (f) is the curve of the number of clusters during pair merging. The merging procedure converges when the cluster number $c$ coincides with the real one of clusters.

\subsubsection{Aggregation to Cores}
Let the resulting clusters of cores be denoted by $\mathcal{C}^t_{\text{core}}=\{\mathcal{C}_1,\dots,\mathcal{C}_c\}$. We need to assign the remaining data points in $\mathcal{C}_{\text{cluster}}\setminus\mathcal{C}^t_{\text{core}}$ to $\mathcal{C}^t_{\text{core}}$. We propose applying the leave-one-out strategy for assignment. The structural affinity of a point $\boldsymbol{x}_i$ to a cluster $\mathcal{C}_j$ can be quantized by the variational value of its rank if we leave $\mathcal{C}_j$ out from $\mathcal{C}_{\text{cluster}}$. For our framework, the ranks of $\boldsymbol{x}_i$ are in-degrees and out-degrees of order $t$. Therefore, we can investigate the ratio of $d^{in}_{i|\{\mathcal{C}_{\text{cluster}}\setminus \mathcal{C}_j\}}$ to $d^{in}_{i|\{\mathcal{C}_{\text{cluster}}\}}$, where the general expression $d_{i|\{\mathcal{C}\}}$ means the degree rank of $\boldsymbol{x}_i$ on $\mathcal{C}$. We compute the same ratio for $d^{out}_i$. Putting these two dual ranks together, we derive the similarity measure of point-to-cluster affinity by product, writing it as%\vspace{-0.1cm}
\begin{equation}
\rho_{\boldsymbol{x}_i\rightarrow \mathcal{C}_j} = 1-\frac{d^{in}_{i|\{\mathcal{C}_{\text{cluster}}\setminus \mathcal{C}_j\}}}{d^{in}_{i|\{\mathcal{C}_{\text{cluster}}\}}}  \frac{d^{out}_{i|\{\mathcal{C}_{\text{cluster}}\setminus \mathcal{C}_j\}}}{d^{out}_{i|\{\mathcal{C}_{\text{cluster}}\}}}=1-\frac{\gamma_{i|\{\mathcal{C}_{\text{cluster}}\setminus \mathcal{C}_j\}}}{\gamma_{{i|\{\mathcal{C}_{\text{cluster}}\}}}},%\vspace{-0.1cm}
\end{equation}
where $\gamma_{{i|\mathcal{C}}}=d^{in}_{i|\mathcal{C}}d^{out}_{i|\mathcal{C}}$ is the product rank of $\boldsymbol{x}_i$.
The larger the value of $\rho_{\boldsymbol{x}_i\rightarrow \mathcal{C}_j}$ is, the more preference $\boldsymbol{x}_i$ has of being attached to $\mathcal{C}_j$. Therefore, the cluster label of $\boldsymbol{x}_i$ can be inferred by%\vspace{-0.1cm}
\begin{equation}\label{eq:aggregate}
 \arg\max_{j}\rho_{\boldsymbol{x}_i\rightarrow \mathcal{C}_j},j=1,\dots,c.%\vspace{-0.2cm}
\end{equation}
Another benefit of applying the ratio of dual degrees to define $\rho_{\boldsymbol{x}_i\rightarrow \mathcal{C}_j}$  is that the ratio can diminish the negative effect of inferring similarity caused by large degrees. The result of attaching toy data to cluster cores is shown in Figure~\ref{fig:toyClusterLayer} (d). To see the correspondence between clusters and homophilic layers, we colorize the HI figure according to the labels of the detected clusters and depict it in Figure~\ref{fig:toyClusterLayer} (g).

\subsubsection{Complexity}
It is straightforward to know that the space complexity of homophilic clustering is $\mathcal{O}(nk_c)$. In practice, $k_c$ is a small integer in $\{1,2,3,4,5\}$. Thus, the complexity reduces to a linear one of $\mathcal{O}(n)$. The time complexity of homophilic clustering depends on the cluster structures of data points. Assume that the maximum number of clusters is $c_{\text{max}}$ during pair merging and the corresponding number of iterations is $m_{\text{max}}$. $c_{\text{max}}$ is actually determined by the connectivity of digraph $\mathcal{G}$ and $k_c$. The time complexity is $\mathcal{O}\left(c_{\text{max}}m_{\text{max}}+(c_{\text{max}}-c)(n_ck_c-m_{\text{max}})\right)$, where $n_c=|\mathcal{C}^t_{\text{core}}|$ is the number of data points in the extracted core clusters. If the given number $c$ of cluster is large, $c$ approaches $c_{\text{max}}$, the complexity will be approximately $\mathcal{O}(cm_{\text{max}})$. If $c$ is small, the worst case will be $\mathcal{O}(n_cc_{\text{max}})$. Usually, $n_c$ is a small fraction of $n$. $c_{\text{max}}$ will be reached for a moderate number $m_{\text{max}}$ of iterations. The curve of $c_{\text{max}}$ over $m_{\text{max}}$ on toy example is shown in Figure~\ref{fig:toyClusterLayer} (f) in the case of $k_c=5$.%\vspace{-0.2cm}
\begin{figure*}%[h]
%\vskip 0.2in
\begin{center}
\begin{tabular}{ccc}
\includegraphics[scale=0.6]{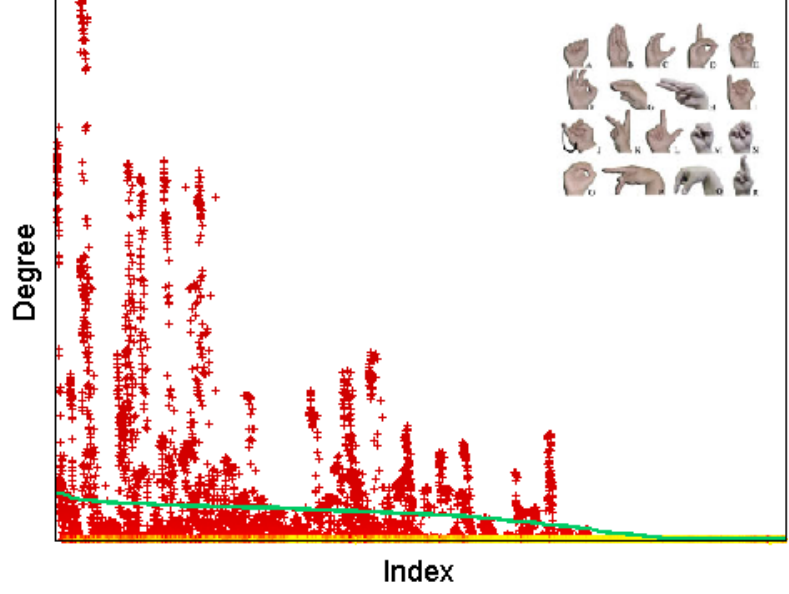}
&
\includegraphics[scale=0.6]{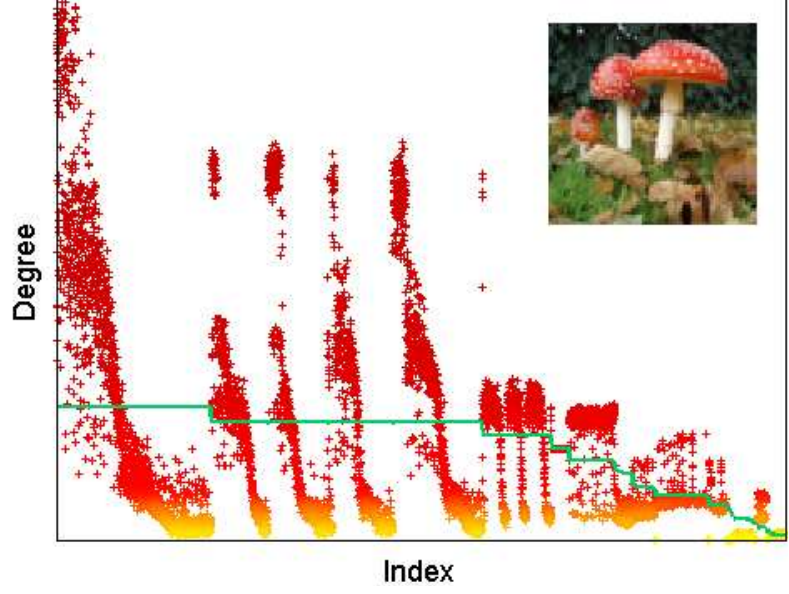}
&
\includegraphics[scale=0.6]{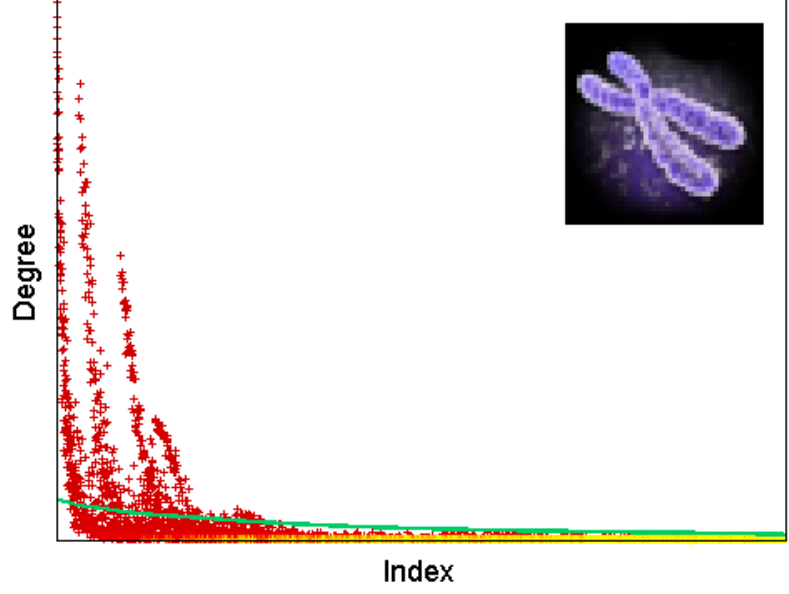}\\
(a) Sign language. & (b) Mushroom. & (c) Gene expression. \\
\includegraphics[scale=0.6]{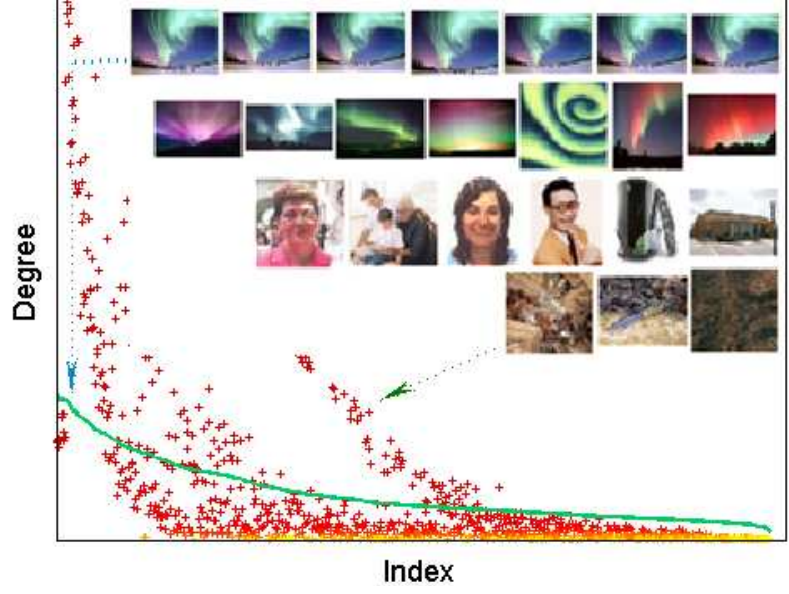}
&
\includegraphics[scale=0.6]{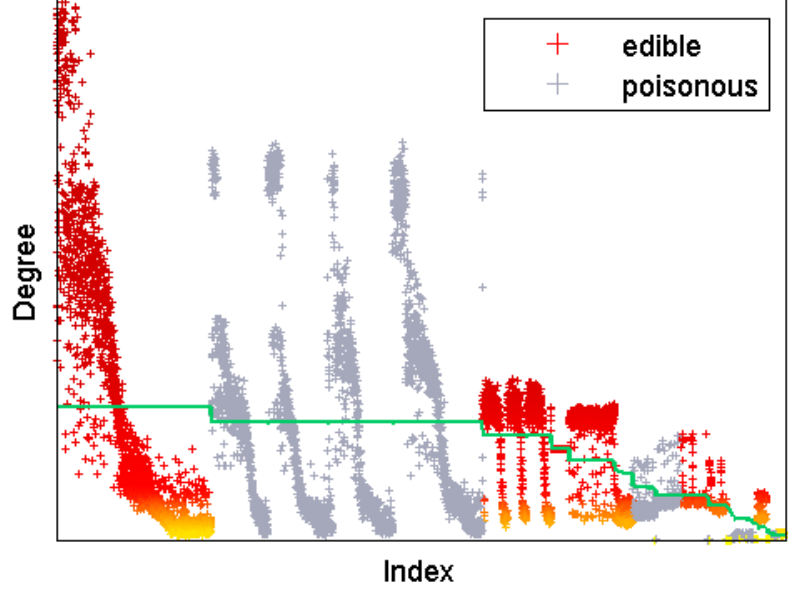}
&
\includegraphics[scale=0.6]{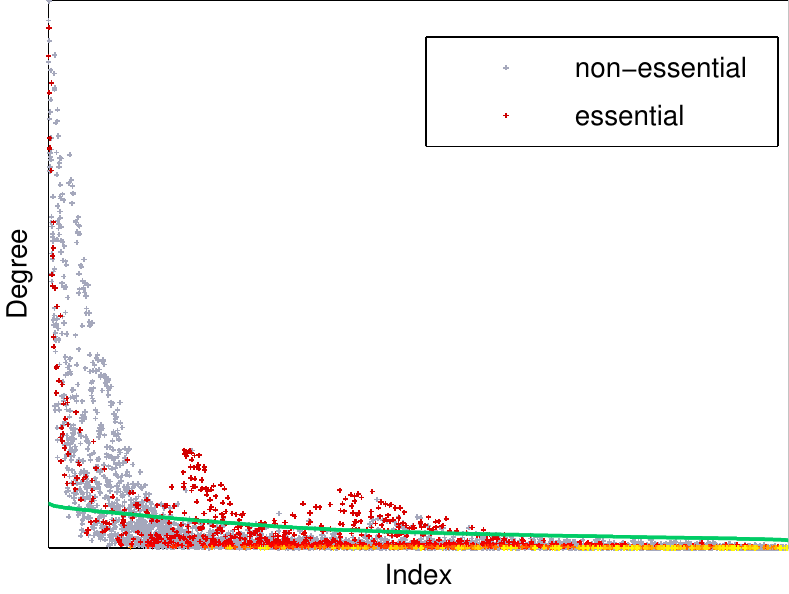}\\
(d) Image retrieval. & (e) Mushroom classification. & (f) Essential genes. \\
\end{tabular}
\end{center}%\vspace{-0.5cm}
\caption{ The HI figures on real-world scientific data. The optimal $t$ for each data set is determined by the approach presented in Section~\ref{section:boundary}. For the figure of gene data in (c), the optimal $t$ is $45$. For the figure of showing essential genes in (f), we carefully select the value of $t$ to be $13$ to highlight the homophilic layer of essential genes.     }
\label{fig:realClusterLayer}
%\vskip -0.0in
\end{figure*}

\section{Experiment}
\label{experiment}
We present more examples of the HI figures and compare our homophilic clustering algorithm with state-of-art algorithms on real-world data. The specific information of each data set is described in the supplementary material paper, which also contains the corresponding graph construction.

\subsection{HI Figure}
The homophilic effect of high-order in-degrees is also observed in real-world data from various scientific domains, as Figure~\ref{fig:realClusterLayer} shows. Figure~\ref{fig:realClusterLayer} (a) shows that the cluster densities of the hand-sign language data are very complicated, forming many homophilic layers. For Web images retrieved by the search engine, the clusters with clear semantics are detected in the HI layers, as Figure~\ref{fig:realClusterLayer} (d) depicts. The semantically meaning images are contained in the strong HI layer, and the noisy images fall into the weakest HI layer and the base of the HI figure. An interesting observation is that many Web images of the same contextual content in the equal or different sizes are solely segregated to be a small agglomerate layer, which is presented by the vertical dotted arrow. This suggests the automatic filtering of redundant information for content-based image retrieval, which plays a central role in the next generation of search engines. For the machine-intelligent discrimination of edible mushrooms from poisonous ones, the edible patterns exhibit transparently-layered regularity, providing considerable ease of classification, as shown by Figures~\ref{fig:realClusterLayer} (b) and (e).
\begin{table*}[t]
\caption{ Performance of clustering (\%). The accuracy of clustering is measured by Normalized Mutual Information (NMI) \cite{NMI}. The abbreviations of the involved algorithms are listed as follows. Linkage, the average linkage algorithm. Zell, Zeta merging based on local links \cite{Zhao08}. Ncuts, Normalized Cuts \cite{Shi00}. SCK, spectral clustering with k-means \cite{Ng01}. MCL, Markov Clustering \cite{MCL00}. AS, Authority shifting \cite{Cho10}. HC, Homophilic Clustering (this paper). The `attribute' refers to the most important feature of the category that the associated algorithm falls into. The denotation `-' represents that the corresponding algorithms are computationally in-feasible for the data set.}
\label{table:clustering}
%\vskip 0.15in
\begin{center}
%\begin{tabular}{|l|c|c|c|c|c|c|c|c|}
\begin{tabular}{lcccccccc}
\hline
Algorithmic Attribute & \multicolumn{1}{c}{Partitional} & \multicolumn{2}{c}{Agglomerative}  & \multicolumn{2}{c}{Spectral} & \multicolumn{3}{c}{Matrix power}\\
Algorithm & k-means  & Average Linkage  &  Zell  & Ncuts  & SCK  & MCL & AS & \textbf{HC}\\
%\hline\hline
\hline
%\\ \hline \\
FRGC & 90.4    & 95      &  \textbf{98.1}  &   92.4 &  90.7 &  88.2  &   88.9   &  97.3 \\
COIL & 82.4   &  89.5   &  91   &   81.9 & 80.5  &  82.3  &   82.9      &  \textbf{97.2} \\
%USPS & 46.1   &  68.6  &     79.9    &   64.7   &  65.4  &  70.2  &   56.6   &  \textbf{85.6}  \\
MNIST & 54.6  & -    &  - &   63.9  & 66.9 &  -  &   -   &  \textbf{81.4} \\
\hline
\end{tabular}
\end{center}
\vskip -0.0in
\end{table*}

Of special scientific interest is the intriguing phenomenon observed from the kinetics of the HI layers of gene expressions in the budding yeast, \emph{Saccharomyces cerevisiae} (Figure \ref{fig:realClusterLayer} (c)).  In network biology, there has been lively debate in recent years concerning the spatial distribution of essential genes in functional modules of networks \cite{Barabasi11}. In light of our findings, a more elaborate structural organization of genes can be revealed from the HI figure. We have carefully checked the growth of the HI figure, and found a meaningful HI layer about essential genes around $t=13$. Figure~\ref{fig:realClusterLayer} (f) illustrates that only a small fraction of essential genes lie in the strong hub (core cluster) with the majority being peripheral. Interestingly, there are the two weak HI layers in which essential genes massively dominate. These two layers are so weak that they rapidly decay with the growth of the HI layers. This observation contributes to evidence that genes possess functional modules that are substantially composed of essential genes, and the sub-networks associated with these modules are very vulnerable,  which is evident because the essential layers transiently exist. In addition, a considerable number of essential genes live in the base of the HI figure, implying that they are dispersively distributed outside functional communities. It is worth noting that these details are apparent only in the moderate evolution of dual degrees over $t$. Video 2 shows the complete dynamic process of the growth of the HI figure.

These examples verify that the HI figure can capture the intrinsic structures of data and is a powerful tool for data visualization and analysis.

\subsection{Clustering}
We perform the experiments of pattern clustering on the three widely applied benchmark databases in face recognition, object classification, and handwritten digit recognition. The face data are from the FRGC (Face Recognition Grand Challenge) database\footnote{http://www.frvt.org/FRGC/}, which contains 466 persons (clusters) of 16,028 facial images. The number of members in each cluster varies from 2 to 80.  The data set of object classification is the processed COIL database\footnote{http://www.cs.columbia.edu/CAVE/software/softlib/coil-100.php}, which contains 7,200 images of 100 objects. Each object cluster has 72 imagery members. The handwritten digits are from the well-known MNIST database\footnote{http://yann.lecun.com/exdb/mnist/}. The MNIST data set includes 70,000 handwritten digits of 10 classes. The algorithms we select to compare are presentative for clustering and most relevant to ours. For graph-based algorithms, we adopt the same directed graph for all algorithms, which can guarantee the fair comparison. We list the compared algorithms and the accuracy of each algorithm in Table~\ref{table:clustering}.

\begin{figure}[t]
\vskip 0.in
\begin{center}
   \includegraphics[scale=0.8]{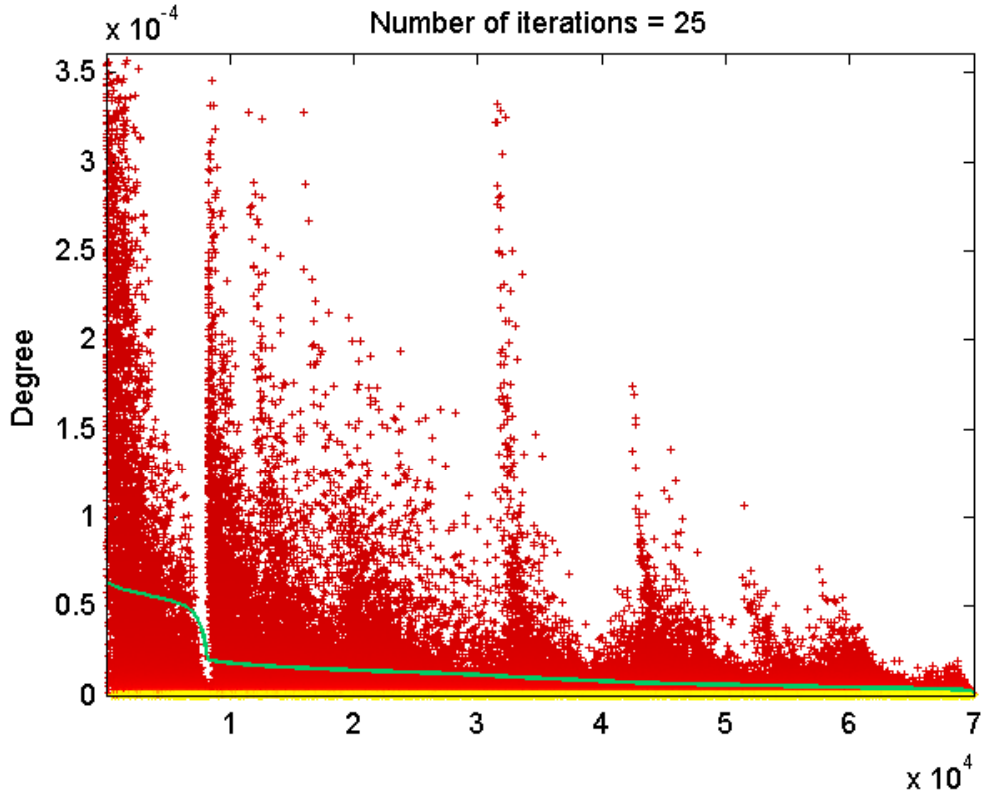}\\
\end{center}%\vskip -0.2in
   \caption{The HI figure on MNIST data $(t=25)$. For good visualization, we truncate the top part of the first layer to highlight the shapes of the remaining layers.}%\vspace{-0.5cm}
\label{fig:MNIST_layer}
%\vskip -0.1in
\end{figure}

Table~\ref{table:clustering} shows that on the relatively simple data like FRGC, the graph-theoretic methods based on hierarchical agglomerative clustering yield the best results and our HC performs comparably well. With the complexity of data increasing, the superiority of HC emerges. On the COIL data, HC is considerably better than the remaining algorithms. On MNIST, our algorithm significantly outperforms all the compared algorithms.The result of clustering the MNIST data proves the robustness of our algorithm to noisy data. Those algorithms of space complexity $\mathcal{O}(n^2)$ are computationally prohibitive for the 70,000-scale MNIST data. The HI figure of MNIST data is shown in Figure~\ref{fig:MNIST_layer} for interested readers' reference. Note that the single linkage algorithm can be scaled to cluster the 70,000 MNIST data. However, its accuracy on MNIST is too low and much lower than the average linkage algorithm on the other three datasets. So we show the results of the average linkage.
%-----------------------------------

\section{Conclusion}
\label{conclusion}
We have reported an interesting property of geometric digraph drawn from neighborhood asymmetries of data. The similarity propagation of local asymmetries leads to the homophilic distribution of in-degrees. Based on this finding, we have proposed an approach called the homophilic in-degree figure to data visualization and developed an algorithm to detect clusters from heavy noise. Extensive experiments on toy data and real scientific data validated the effectiveness of our algorithms. In addition to the applications in pattern clustering, our algorithms can also be applicable for vector quantization, Nystr\"{o}m matrix approximation, topic models, and image segmentation, in which cases clusters play an important role.

\section*{Acknowledgement}
We are aware that a paper published on Science very recently \cite{Rodriguez14} handles the similar problem of clustering with the one presented in this paper.

\bibliography{HC}
\bibliographystyle{icml2014}

\end{document}